\documentclass[10pt,twocolumn,letterpaper]{article}

\usepackage{cvpr}
\usepackage{times}
\usepackage{epsfig}
\usepackage{graphicx}
\usepackage{amsmath}
\usepackage{amssymb}

\usepackage{bm}
\usepackage{color}
\usepackage{subfig}

\usepackage{makecell}
\usepackage{url}
\usepackage{multirow}
\usepackage{footnote}

\makeatletter
\renewcommand{\maketag@@@}[1]{\hbox{\m@th\normalsize\normalfont#1}}%
\makeatother

\usepackage[breaklinks=true,bookmarks=false]{hyperref}

\cvprfinalcopy 

\def\cvprPaperID{****} 

\ifcvprfinal\fi
\begin{document}

\title{Minimal Solvers for Mini-Loop Closures in 3D Multi-Scan Alignment}

\author{Pedro Miraldo\\
KTH Royal Institute of Technology\\
{\tt miraldo@kth.se}
\and
Surojit Saha\ and\ Srikumar Ramalingam\\
School of Computing, The University of Utah \\
{\tt \{surojit,srikumar\}@cs.utah.edu}
}

\maketitle
\thispagestyle{empty}

\begin{abstract}
   3D scan registration is a classical, yet a highly useful problem in the context of 3D sensors such as Kinect and Velodyne. While there are several existing methods, the techniques are usually incremental where adjacent scans are registered first to obtain the initial poses, followed by motion averaging and bundle-adjustment refinement. In this paper, we take a different approach and develop minimal solvers for jointly computing the initial poses of cameras in small loops such as 3-, 4-, and 5-cycles\footnote{A cycle graph ${\cal C}_n$, also referred to as $n$-cycles, is a subgraph with $n$ nodes and edge set $\{(1,2),\dots,(n-1,n),(n,1)\}$.}. Note that the classical registration of 2 scans can be done using a minimum of 3 point matches to compute 6 degrees of relative motion. On the other hand, to jointly compute the 3D registrations in $n$-cycles, we take 2 point matches between the first $n-1$ consecutive pairs (i.e., Scan 1 \& Scan 2, $\dots$, and Scan $n-1$ \& Scan $n$) and 1 or 2 point matches between Scan $1$ and Scan $n$. Overall, we use 5, 7, and 10 point matches for 3-, 4-, and 5-cycles, and recover 12, 18, and 24 degrees of transformation variables, respectively. Using simulations and real-data we show that the 3D registration using mini $n$-cycles are computationally efficient, and can provide alternate and better initial poses compared to standard pairwise methods.
\end{abstract}

\section{Introduction}

Many geometers working on algebraic minimal solvers have attempted to solve the notorious and classical 3-view 4-point relative pose estimation. Given 4 triplets of point matches, the goal is to jointly find the poses of the 3 cameras. There have been some great progress on this problem using one-dimensional search~\cite{Nister2006} and semi-definite programming~\cite{Li2010}, but we still miss the simple and direct minimal algebraic solver that we usually derive for most geometric vision problems. If one manages to solve this problem for RGB cameras, what would be the next big challenge? Do we look at the 4-view 3-point relative pose problem? While there has been a great deal of effort to solve the higher-order pose estimation in the case of RGB sensors, the equivalent problem with RGB-D cameras has received no attention. In the case of RGB-D sensors, the number of correspondences for the $n$-camera relative pose problem is less notorious for $n \leq 5$, and even practically deployable. At this point, when the price point for commercial RGB-D sensors is decreasing due to the progress in robotics and self-driving industry, it would be a good time to fully equip the arsenal with algebraic minimal solvers for depth sensors. 

\begin{figure}
    \centering
    \includegraphics[width=0.47\textwidth]{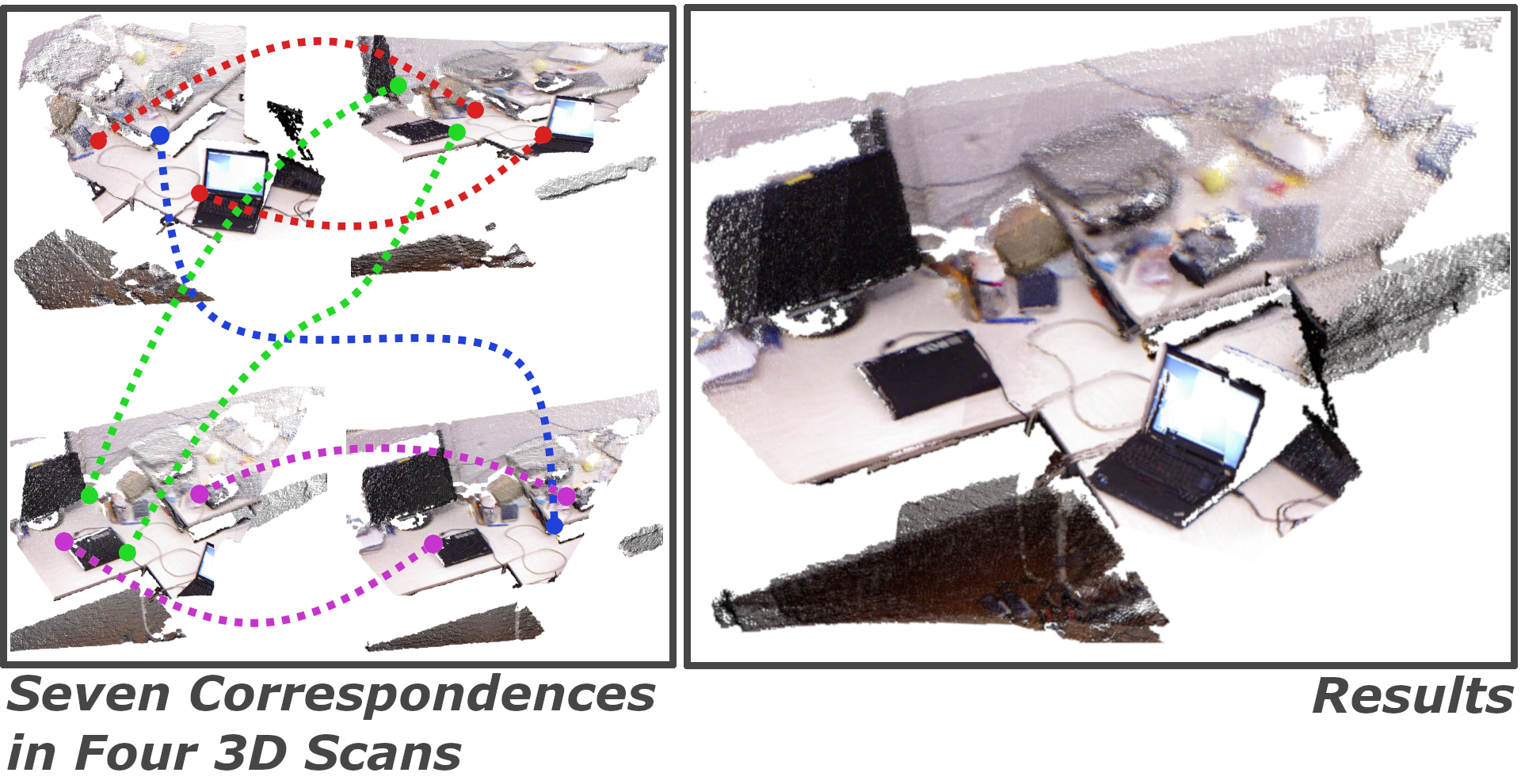}
    \caption{\it At the left we show four scans and seven 3D point matches (2 from Scans 1 \& Scan 2; 2 from Scans 2 \& Scan 3; 2 from Scans 3 \& Scan 4; and 1 from Scans 1 \& Scan 4). At the right, we show the registered scan using our minimal solver for 4-cycle, that jointly computes the pose parameters for all the four cameras.}
    \label{fig:intro}
\end{figure}

In Fig.~\ref{fig:intro} we show four different scans collected using a Kinect sensor. We jointly compute the 3D registration for all four scans using a minimal solver that uses a total of seven point matches. We are able to compute 18 degrees of transformation variables, and the points from all the four scans are registered as shown in Fig.~\ref{fig:intro}. Previous methods for RGB-D registration typically employ pair-wise registration where the initial poses are computed between pairs of cameras, and a final refinement is done using a non-linear refinement technique. The pairwise methods (see Orthogonal Procrustes problem \cite{schonemann66} that uses a minimum of three point matches) typically accumulate drift even in the case of three cameras. Our formulation naturally eliminates the drift in these mini $n-$cycles, and thereby provides better pose parameters. This paper systematically studies the possibility of joint 3D registration for mini $n-$cycles, and derives algebraic minimal solvers, which are typically embedded in a Random Sample Consensus (RANSAC)~\cite{fischler1981} framework for robust estimation of pose parameters. It has been well established that minimal solvers and RANSAC tend to perform robustly in the presence of outliers.

\subsection{Related Work}
We carefully survey some of the classical and modern registration algorithms that employ 3D sensors. 

\vspace{0.25cm}\noindent{\bf 3D scan alignment:~}
The classic approach to solve the 3D scan alignment problem is the Iterated Closest Point (ICP) algorithm, proposed in \cite{besl92}. Over the years, several efficient and robust solutions have been proposed in the literature to solve the 3D multi-scan alignment using 3D points, such as \cite{segal09,myronenko10,newcombe11,yang13,venu14,li17,park17}. A method for fast and efficient 3D rotation search is proposed in ~\cite{bustos16}.

Besides the classic approaches presented above, alternate methods have been proposed that utilize the properties of the observed 3D scene. In \cite{endres14}, a beam-based environment measurement model was introduced to achieve frame-to-frame registration. In \cite{Ramalingam_2013,ma16,bhattacharya17,liu18} we use 3D planes to improve the SLAM using 3D cameras. In \cite{lu15} we extract and use 3D straight lines for 3D SLAM, while \cite{choi13} focuses on edge detection. In \cite{halber17}, a more general method is proposed to detect and enforce constraints (geometric relationships and feature correspondences). Surveys on the evaluation of 3D SLAM methods were presented in~\cite{endres12,sturm12}.
There have also been some solvers for the non-rigid 3D registration problems (see for example \cite{zollhofer12,bernard17,slavcheva17,ma17}). A survey on rigid and non-rigid registration of 3D point clouds is presented in \cite{tam13}.

In addition to finding the 3D transformations that align 3D scans, there have been some developments on doing both the 3D registration and semantic segmentation using RGB-D images. Several works were proposed such as \cite{tatarchenko18,schonberger18,zaganidis18,zeng18}.

Recently, some deep learning techniques techniques were used in order to obtain 3D registration. In \cite{elbaz17}, local 3D geometric structures are extracted using a deep neural network auto-encoder. Compact geometric features are learned in \cite{khoury17}. Automatic reconstruction of floorplans is achieved using a deep learning algorithm in \cite{liu18}.

\vspace{0.25cm}\noindent{\bf Minimal solvers:~}
We review some of the minimal solutions that are relevant to pose estimation using RGB cameras. Several solutions were proposed for the absolute pose for central perspective cameras (three 3D point correspondences between the world and image), see for example~\cite{kneip11,ke17,wang18,persson18}. The pose estimation has also been studied for the pose of multi-perspective systems, such as~\cite{ventura14,gim16,camposeco16,miraldo18}.

When considering the relative pose estimation, several approaches have also been proposed for solving the minimal relative pose problem. See for example \cite{mister04,li06b} for calibrated cameras. There are other solutions such as \cite{li13} which studies the relative pose estimation with a known relative rotation angle, \cite{fraundorfer10,saurer15} for the relative pose with known directions, \cite{li06} for the relative pose with unknown focal length, solutions invariant to translation \cite{kneip12}, and solutions to the generalized relative pose problem \cite{stewenius05c,ventura15}. In \cite{camposeco18}, a hybrid minimal solver that combines relative with absolute poses is proposed.

\subsection{Notation and Problem Definition}\label{sec:prob_definition}
For simplicity, we use ${\cal S}_n$ to denote Scan $n$. The $i$\textsuperscript{th} 3D point in ${\cal S}_n$ is denoted as $\mathbf{p}_i^{{\cal S}_n}\in\mathbb{R}^4$, which is represented in homogeneous coordinates. Rotation matrices and translation vectors are denoted as $\mathbf{R}^{{\cal S}_n,{\cal S}_m}\in\mathcal{SO}(3)$ \& $\mathbf{t}^{{\cal S}_n,{\cal S}_m}\in\mathbb{R}^3$, for transformations from ${\cal S}_n$ to ${\cal S}_m$. We use the $n$-cycle to denote the sequences of $n$ 3D scans with loop closure (first and last point clouds in the sequence have 3D point correspondences).

The goal is to find the transformation matrices $\mathbf{T}^{{\cal S}_n,{\cal S}_m}\in\mathcal{SE}(3)$ that transform 3D points from coordinate system ${\cal S}_n$ to ${\cal S}_m$ such that
\begin{equation}
    \mathbf{p}_i^{{\cal S}_m}
    \simeq
    \underbrace{\begin{bmatrix}
    \mathbf{R}^{{\cal S}_n,{\cal S}_m} & \mathbf{t}^{{\cal S}_n,{\cal S}_m} \\
    \mathbf{0}_{1,3} & 1 
    \end{bmatrix}}_{\mathbf{T}^{{\cal S}_n,{\cal S}_m}}
    \mathbf{p}_i^{{\cal S}_n}.
\end{equation} 
We are given sets of 3D point matches $(\mathbf{p}_i^{\mathcal{S}_n}, \mathbf{p}_i^{\mathcal{S}_m})$. Symbol $\simeq$ denotes that the terms are equal up to a scale factor.

\begin{table}
\centering
\scalebox{0.73}{
\begin{tabular}{ |c|l|c|c| }
 \hline
 \bf \thead{Cycle \\ \#Cameras} & \multicolumn{1}{c|}{\bf \#Correspondences}  & \bf Total & \bf \#Solutions \\
 \hline\hline
 Two   & $\#\bm{3}({\cal S}_1,{\cal S}_2)$  & 3 & 2 \\ \hline
 Three & $\#\bm{2}({\cal S}_1,{\cal S}_2); \ \#\bm{2}({\cal S}_2,{\cal S}_3);\ \#\bm{1}({\cal S}_1,{\cal S}_3);$  & 5 & 4 \\ \hline
 Four  & $\begin{aligned} & \#\bm{2}({\cal S}_1,{\cal S}_2); \ \#\bm{2}({\cal S}_2,{\cal S}_3); \ \#\bm{2}({\cal S}_3,{\cal S}_4); & \\ &  \#\bm{1}({\cal S}_1,{\cal S}_4) & \end{aligned} $ & 7 & 16 \\ \hline
 Five  & $\begin{aligned} & \#\bm{2}({\cal S}_1,{\cal S}_2); \ \#\bm{2}({\cal S}_2,{\cal S}_3); &  \\ & \#\bm{2}({\cal S}_3,{\cal S}_4); \ \#\bm{2}({\cal S}_4,{\cal S}_5) \ \#\bm{2}({\cal S}_1,{\cal S}_5) & \end{aligned}$ & 10 & 32 \\ \hline
\end{tabular}}
\caption{\emph{This table summarizes the minimal number of correspondences required to compute the 3D point registration. In the table, $\#\bm{i}({\cal S}_j,{\cal S}_k)$ means $i$ point correspondences within the sequence of point clouds ${\cal S}_j$ and ${\cal S}_k$.}}
\label{tab:sumup3D}
\end{table}

\vspace{.25cm}\noindent
{\bf Contributions:} We propose  novel minimal solvers for the mini $n-$cycles in 3D point cloud registration. We propose three solvers for 3-, 4-, and 5-cycles for the general six degrees of freedom and planar motions. The Tab.~\ref{tab:sumup3D} highlights the different $n-$cycles, required point correspondences, and the number of solutions. To the best of our knowledge, we are the first to propose and solve these cases. 

\section{Minimal Solvers}\label{sec:camera_solvers}

\begin{figure}[!htbp]
\centering
    \subfloat[\emph{Predefined transformations for two point clouds and two correspondences. The remaining degree of freedom is $\alpha$. Note that $\bm{e}$ and $\bm{f}$ are the transformed coordinate frames of ${\cal S}_1$ and ${\cal S}_2$, respectively.}]{
     \includegraphics[width=0.98\linewidth]{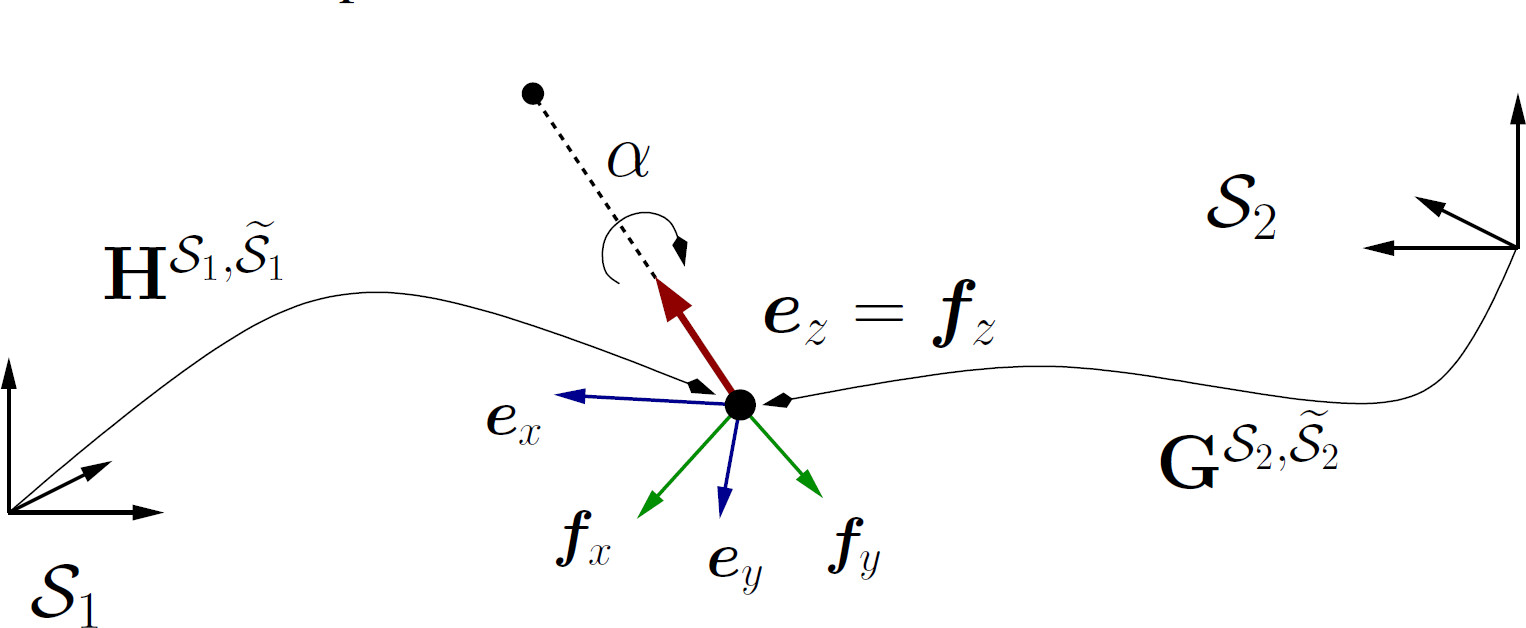}
	\label{fig:predefined_transformations}
     } \\
     \subfloat[\emph{Three point clouds, two correspondences between ${\cal S}_1$ \& ${\cal S}_2$ and ${\cal S}_2$ \& ${\cal S}_3$. $\alpha$ and $\beta$ are the remaining degrees of freedom.}]{
     \includegraphics[width=0.98\linewidth]{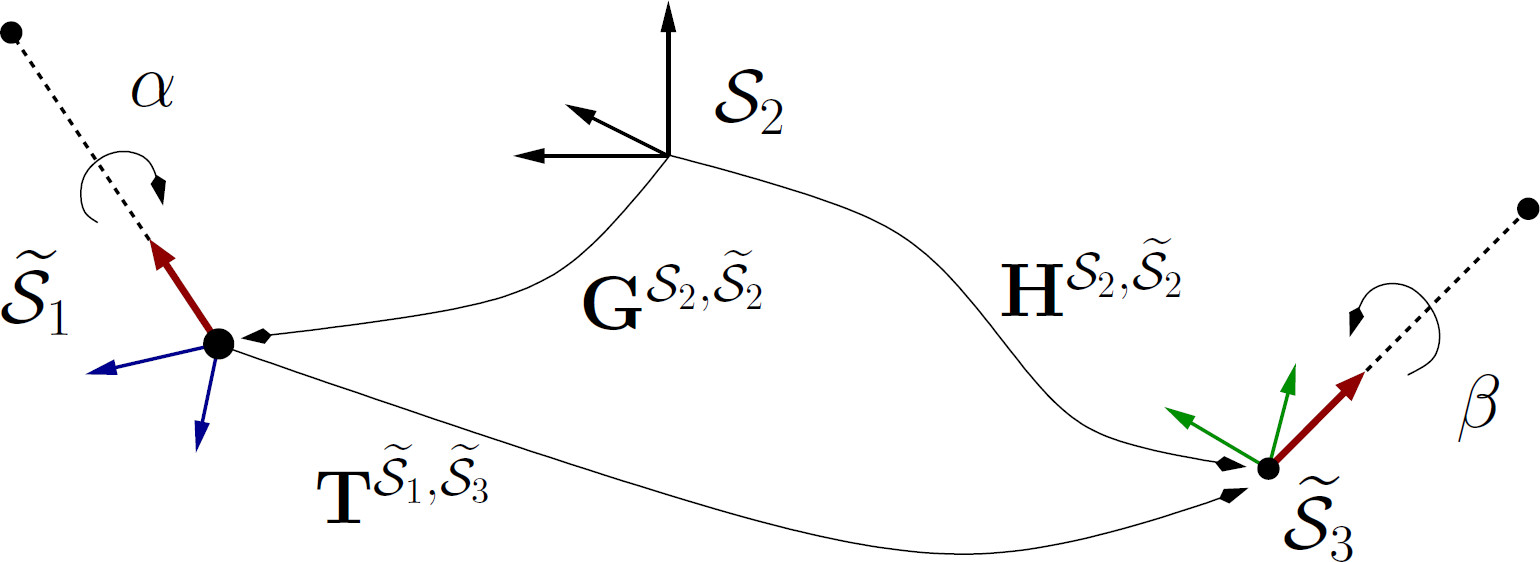}
	\label{fig:result_dof}
     } \\
    \subfloat[\emph{Four point clouds, two correspondences between ${\cal S}_1$ \& ${\cal S}_2$, ${\cal S}_2$ \& ${\cal S}_3$, and ${\cal S}_3$ \& ${\cal S}_4$. $\alpha$, $\beta$, and $\gamma$ are the remaining degrees of freedom.}]{
    \includegraphics[width=0.98\linewidth]{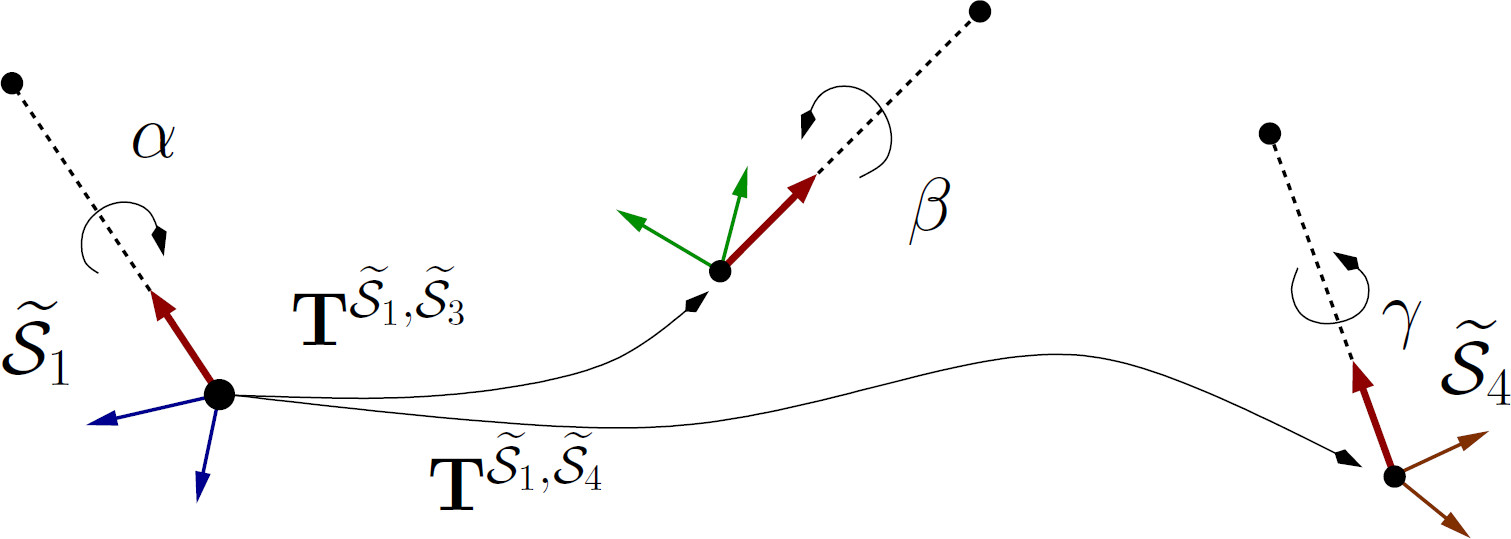}
    \label{fig:result_dof_four}
    }
    \caption{\emph{Representation of the predefined transformations \protect\subref{fig:predefined_transformations} and the resulting degrees of freedom for the three camera 3D registrations, with two point correspondences between point clouds one \& two and two \& three \protect\subref{fig:result_dof}. \protect\subref{fig:result_dof_four} shows the remaining degrees of freedom for four point clouds and two 3D point correspondences between one \& two, two \& three, and three \& four.}}
    \label{fig:prob_simp}
\end{figure}

In this section, we formulate the minimal solution for jointly estimating the poses of $n-$cameras that occur in an $n-$cycle. In all the $n-$cycles, when $n > 2$ we use a simple geometric idea. Let us assume that we would like to find the registration between two different camera scans ${\cal S}_1$ and ${\cal S}_2$. As shown in Fig.~\ref{fig:prob_simp}\subref{fig:predefined_transformations}, the basic idea is to first use two point correspondences to construct a virtual axis passing through these two points. Now we align the coordinate frames of ${\cal S}_1$ and ${\cal S}_2$ in such a manner that the $z-$axis of both these frames are aligned along this virtual axis. The triplets $\{\bm{e}_x,\bm{e}_y,\bm{e}_z\}$ and $\{\bm{f}_x,\bm{f}_y,\bm{f}_z\}$ denote the coordinate frames for both these cameras after the alignment. Next, the problem of estimating the transformations between these coordinate frames can be seen as just estimating the rotation angle around the $z-$axis. This idea of using simple predefined transformations before the actual registration allows us to simplify the constraint equations. Once we obtain the final registration, we can always find the relative poses between the original coordinate frames, by just using the inverses of the predefined transformation matrices.

Next, we show the details of the predefined transformations that we use on the original scans, so that the actual minimal solvers become easier to derive (see Tab.~\ref{tab:sumup3D}).

\subsection{Setting the Stage for Minimal Solvers}\label{sec:get_dof}

Let us consider two point matches $(\mathbf{p}_{1}^{{\cal S}_1},\mathbf{p}_{2}^{{\cal S}_1})$ and $(\mathbf{p}_{1}^{{\cal S}_2},\mathbf{p}_{2}^{{\cal S}_2})$ in ${\cal S}_1$ and ${\cal S}_2$, respectively.
We consider the predefined transformations to align the scans such that the new coordinates frames of ${\cal S}_1$ and ${\cal S}_2$ satisfy the following conditions:
\begin{itemize}
    \item Centered in $\mathbf{p}_{1}^{{\cal S}_1}$ and $\mathbf{p}_{1}^{{\cal S}_2}$, respectively;
    \item $z-$axis of both frames are aligned with directions $(\mathbf{p}_{2}^{{\cal S}_1}$ - $\mathbf{p}_{1}^{{\cal S}_1})$ and $(\mathbf{p}_{2}^{{\cal S}_2}$ - $\mathbf{p}_{1}^{{\cal S}_2})$, respectively.
\end{itemize}
A depiction of these predefined transformations is shown in Fig.~\ref{fig:prob_simp}\subref{fig:predefined_transformations}.
To get these, we define transformation matrices $\mathbf{H}^{{\cal S}_1,\widetilde{\cal S}_1},\mathbf{G}^{{\cal S}_2,\widetilde{\cal S}_2}\in\mathcal{SE}(3)$ such that
\begin{equation}
    \label{eq:simp_pointcloud}
    \mathbf{p}_{1,2}^{\widetilde{\cal S}_1}
    \simeq
    \mathbf{H}^{{\cal S}_1,\widetilde{\cal S}_1}
    \mathbf{p}_{1,2}^{{\cal S}_1} \ \ \  \text{and} \ \ \
    \mathbf{p}_{1,2}^{\widetilde{\cal S}_2}
    \simeq
    \mathbf{G}^{{\cal S}_2,\widetilde{\cal S}_2}
    \mathbf{p}_{1,2}^{{\cal S}_2},
\end{equation}
where $\widetilde{\cal S}_n$ denotes the transformed point clouds and
\begin{align}
    \label{eq:predefined_trans_1}
    \mathbf{H}^{{\cal S}_1,\widetilde{\cal S}_1} = &
    \begin{bmatrix}
    \mathbf{U}^{{\cal S}_1,\widetilde{\cal S}_1} & \mathbf{0} \\
    \mathbf{0} & 1 
    \end{bmatrix}
    \begin{bmatrix}
    \mathbf{I} & -\mathbf{q}_1^{{\cal S}_1} \\
    \mathbf{0} & 1
    \end{bmatrix} \ \text{and} \\
    \label{eq:predefined_trans_2}
    \mathbf{G}^{{\cal S}_2,\widetilde{\cal S}_2} = &
    \begin{bmatrix}
    \mathbf{V}^{{\cal S}_2,\widetilde{\cal S}_2} & \mathbf{0} \\
    \mathbf{0} & 1 
    \end{bmatrix}
    \begin{bmatrix}
    \mathbf{I} & -\mathbf{q}_1^{{\cal S}_2}\\
    \mathbf{0} & 1
    \end{bmatrix},
\end{align}
in which $\mathbf{U}^{{\cal S}_1,\widetilde{\cal S}_1}, \mathbf{V}^{{\cal S}_2,\widetilde{\cal S}_2}\in\mathcal{SO}(3)$ are any rotation matrices that align the $z-$axis of ${\cal S}_1$ and ${\cal S}_2$ (respectively) with the direction from $\mathbf{p}_1$ to $\mathbf{p}_2$, and $\mathbf{q}_1\in\mathbb{R}^3$ represents the regular coordinates of $\mathbf{p}_1$.

The transformation matrix from $\mathcal{S}_1$ to $\mathcal{S}_2$, after applying predefined transformations, is as follows:
\begin{equation}
    \label{eq:def_t12}
    \mathbf{T}^{{\cal S}_1,{\cal S}_2} =
    \left.\mathbf{G}^{{\cal S}_2,\widetilde{\cal S}_2}\right.^{-1}\underbrace{
    \begin{bmatrix}
    c\alpha & -s\alpha & 0 & 0 \\
    s\alpha &  c\alpha & 0 & 0\\
    0 & 0 & 1 & 0 \\
    0 & 0 & 0 & 1
    \end{bmatrix}}_{\mathbf{L}(\alpha)} \mathbf{H}^{{\cal S}_1,\widetilde{\cal S}_1},
\end{equation}
where $\mathbf{L}(\alpha)$ is a single degree of freedom transformation matrix representing a rotation around the $z-$axis. We use $c\alpha$ and $s\alpha$ to denote $\text{cos}(\alpha)$ and $\text{sin}(\alpha)$, respectively.

Once we align the coordinate frames using the predefined transformations, all we have to compute is one rotation angle for every pair of 3D scans (see Fig.~\ref{fig:prob_simp}). So, for the case of having two scans, we just focus on getting the one unknown rotation from Scan 1 to Scan 2. In the next few sections, we show the minimal solutions for $n-$cycles. Note that this idea of using virtual axis to register scans is straightforward in the case of two cameras, but a little intriguing when we start using multiple axes. For different pairs of cameras in the case of $n-$cycles, when $n > 2$, the underlying idea is still the same. We use only 2 point correspondences between different pairs of 3D scans to realize the predefined transformations (refer to Fig.~\ref{fig:prob_simp}\subref{fig:predefined_transformations}). Following this, we just need to find the corresponding rotation angles.

\subsection{Pairwise Registration}\label{sec:two_camera_solver}
We show the two camera registration for illustrating the idea. By considering the predefined transformations defined in the previous subsection, this can be easily achieved by considering a third point correspondence between $\widetilde{\cal S}_1$ and $\widetilde{\cal S}_2$ (see \eqref{eq:simp_pointcloud}), and checking for $\alpha$ that satisfies
\begin{equation}
    \label{eq:one_dof}
    \mathbf{p}_{3}^{\widetilde{\cal S}_2} \simeq
    \mathbf{L}(\alpha)
    \mathbf{p}_{3}^{\widetilde{\cal S}_1}.
\end{equation}

Notice that \eqref{eq:one_dof} has two linear equations as a function of $c\alpha$ and $s\alpha$, meaning that we can compute a single solution for both variables, and therefore a single solution for $\alpha$. However, when using noisy data, solutions for $c\alpha$ and $s\alpha$ will not satisfy the trigonometric constraints $c\alpha^2 + s\alpha^2 = 1$.
To avoid this, we consider a single constraint of \eqref{eq:one_dof}, which we solve as a function of $c\alpha$ and replace it in $c\alpha^2 + s\alpha^2 = 1$, which gives up to two solution to the problem. Although this approach gives more than one solution, they ensure $\mathbf{L}(\alpha)$ is a rotation matrix and therefore $\mathbf{T}^{{\cal S}_1,{\cal S}_2}$ is a transformation matrix. In addition, one can remove one of the solutions by back-substituting them in \eqref{eq:one_dof}. As in Procrustes's solver, this can be computed in closed-form.

In the following sections, we show the registration for $n-$cycles for $n > 2$. Note that we establish constraints between different pairs of cameras, but the 3D registration for all the cameras is computed by jointly solving all the equations. In other words, the registration is a higher-order one and not solving different pairwise registrations independently. 

\subsection{3-Cycle Registration} \label{sec:three_camera_solver}
Now, let us consider three point clouds ${\cal S}_1$, ${\cal S}_2$, and ${\cal S}_3$, and two correspondences between ${\cal S}_1$ and ${\cal S}_2$, and two correspondences between ${\cal S}_2$ and ${\cal S}_3$. We start by considering some predefined transformations to the point clouds, to ensure that the respective 3D points satisfy the assumptions of Sec.~\ref{sec:get_dof}. We aim at finding $\widetilde{\cal S}_1$, $\widetilde{\cal S}_2$, and $\widetilde{\cal S}_3$ that allow us to write constraints similar to \eqref{eq:one_dof}. For this purpose, one has to find $\mathbf{H}^{{\cal S}_1,\widetilde{\cal S}_1}$, $\mathbf{G}^{{\cal S}_2,\widetilde{\cal S}_2}$, $\mathbf{H}^{{\cal S}_2,\widetilde{\cal S}_2}$, and $\mathbf{G}^{{\cal S}_3,\widetilde{\cal S}_3}$ similar to the ones in \eqref{eq:predefined_trans_1} and \eqref{eq:predefined_trans_2}, such that
\begin{equation}
\label{eq:trans_predef_3cycle}
    \mathbf{p}_{i}^{\widetilde{\cal S}_1}
    \simeq 
    \mathbf{H}^{{\cal S}_1,\widetilde{\cal S}_1}
    \mathbf{p}_{i}^{{\cal S}_1} \ \ \text{and} \ \
    \mathbf{p}_{j}^{\widetilde{\cal S}_3}
    \simeq 
    \mathbf{G}^{{\cal S}_3,\widetilde{\cal S}_3}
    \mathbf{p}_{j}^{{\cal S}_3}.
\end{equation}
Using these predefined transformations, we define the transformation from $\widetilde{\cal S}_1$ to $\widetilde{\cal S}_3$ as
\begin{equation}
    \label{eq:trans32}
    \mathbf{T}^{\widetilde{\cal S}_1,\widetilde{\cal S}_3} =
    \mathbf{L}(\beta) \underbrace{\mathbf{H}^{{\cal S}_2,\widetilde{\cal S}_2} \left.\mathbf{G}^{{\cal S}_2,\widetilde{\cal S}_2}\right.^{-1}}_{\mathbf{K}_2\in\mathcal{SE}(3)} \mathbf{L}(\alpha).
\end{equation}
By doing this, we reduce the problem of estimating the transformation between three 3D scans to two degrees of freedom (in this case angles $\alpha$ and $\beta$). A graphical representation of this problem is shown in Fig.~\ref{fig:prob_simp}\subref{fig:result_dof}.

Now, to compute the transformations we have to use addition information. Let us consider that we have a correspondence between ${\cal S}_1$ and ${\cal S}_3$, i.e. a correspondence to close the cycle between the first and third cameras (notice that additional correspondences between ${\cal S}_1$ \& ${\cal S}_2$ and ${\cal S}_2$ \& ${\cal S}_3$ can be solved by the method proposed in Sec.~\ref{sec:three_camera_solver}).
Let us denote the correspondence point between ${\cal S}_1$ and ${\cal S}_3$ as $\mathbf{p}_5^{{\cal S}_1}$ and $\mathbf{p}_5^{{\cal S}_3}$, respectively. By applying the predefined transformation to the data as shown in \eqref{eq:trans_predef_3cycle}, and using \eqref{eq:trans32}, we get three constraints of the form
\begin{equation}
    \label{eq:3cam_const1}
    \mathbf{p}_5^{\widetilde{\cal S}_3} \simeq 
    \mathbf{L}(\beta) \mathbf{K}_2 \mathbf{L}(\alpha) \mathbf{p}_5^{\widetilde{\cal S}_1}.
\end{equation}
Notice that we have two unknowns and three constraints in \eqref{eq:3cam_const1}. Therefore, in general, it is possible to find $\alpha$ and $\beta$ with only one point correspondence.

To solve this problem, we use the fact that the third constraint in \eqref{eq:3cam_const1} (i.e. its third row) only depends on the unknown parameter $\alpha$:
\begin{equation}
    \label{eq:3cam_solve1}
    a_1 c\alpha + a_2 s\alpha + a_3 = 0,
\end{equation}
where $a_1,a_2,a_3$ are known coefficients.
On the other hand, if we consider the inverse transformation $\mathbf{T}^{\widetilde{\cal S}_3,\widetilde{\cal S}_1}$:
\begin{equation}
    \label{eq:3cam_const2}
    \mathbf{p}_5^{\widetilde{\cal S}_1} \simeq 
    \mathbf{L}(\alpha)^T \mathbf{K}_2^{-1} \mathbf{L}(\beta)^T \mathbf{p}_5^{\widetilde{\cal S}_3}
\end{equation}
and use, again, the third row of \eqref{eq:3cam_const2}, we get a constraint that only depends on $\beta$:
\begin{equation}
    \label{eq:3cam_solve2}
    a_4 c\beta + a_5 s\beta + a_6 = 0.
\end{equation}

Now, to solve the problem we just have to solve \eqref{eq:3cam_solve1} \& \eqref{eq:3cam_solve2}, using the trigonometric constraints $c\alpha^2 + s\alpha^2 = 1$ \& $c\beta^2 + s\beta^2 = 1$. Note that the unknowns are decoupled, meaning that we can compute them separately. This can be done as follows:
1) we solve \eqref{eq:3cam_solve1} as a function of $c\alpha$;
2) substitute the solution in $c\alpha^2 + s\alpha^2 = 1$ (which gives a two degree polynomial equation in $c\alpha$); and
3) compute the roots of the resulting equation giving up to two solutions to $c\alpha$.
The value for $s\alpha$ is given by choosing one in $\{\pm\sqrt{1-c\alpha^2}\}$ that satisfy \eqref{eq:3cam_solve1}.
This procedure is repeated for the $s\beta$ and $c\beta$, giving two additional solutions for these two unknowns. Since the pairs of solutions for $\alpha$ and $\beta$ are decoupled, we will have up to four valid solutions for our problem (as reported in Tab.~\ref{tab:sumup3D}). Next, we study the four 3D scans case.

\subsection{4-Cycle Registration} \label{sec:four_camera_solver}
Let us consider $4$ point clouds.
Again, assume that we have two correspondences between ${\cal S}_1$ \& ${\cal S}_2$, ${\cal S}_2$ \& ${\cal S}_3$, and ${\cal S}_3$ \& ${\cal S}_4$ (see Fig.~\ref{fig:prob_simp}\subref{fig:result_dof_four}).
By following the same assumptions of previous subsections, we get
$\mathbf{T}^{{\cal S}_1,{\cal S}_2}$ as in \eqref{eq:def_t12},
\begin{align}
     \mathbf{T}^{{\cal S}_2,{\cal S}_3} =  & \left(\mathbf{G}^{{\cal S}_3,\widetilde{\cal S}_3}\right)^{-1} \mathbf{L}(\beta) \mathbf{H}^{{\cal S}_2,\widetilde{\cal S}_2}, \ \text{and} \label{eq:def_t23} \\
    \mathbf{T}^{{\cal S}_3,{\cal S}_4} = & \left(\mathbf{G}^{{\cal S}_4,\widetilde{\cal S}_4}\right)^{-1} \mathbf{L}(\gamma) \mathbf{H}^{{\cal S}_3,\widetilde{\cal S}_3}.\label{eq:def_t34}
\end{align}
The matrices $\mathbf{G}$ and $\mathbf{H}$ are given by applying the method in Sec.~\ref{sec:get_dof}. Therefore, we have only three degrees of freedom remaining to get the relative poses between all the four 3D scans. 
More specifically, angles $\alpha$, $\beta$, and $\gamma$.
A trivial solution to this problem would be to consider additional correspondences between ${\cal S}_1$ \& ${\cal S}_2$, ${\cal S}_2$ \& ${\cal S}_3$, or ${\cal S}_3$ \& ${\cal S}_4$.
One could use a combination of the methods presented in the previous subsections to solve the relative positions between the cameras.
However, here we are interested in the 4-cycles, i.e. only one correspondence between ${\cal S}_1$ and ${\cal S}_4$ in addition to the pairwise correspondences.

By premultiplying the transformations defined in \eqref{eq:def_t12}, \eqref{eq:def_t23}, and \eqref{eq:def_t23}, we can define
\begin{equation}
    \mathbf{T}^{\widetilde{\cal S}_1,\widetilde{\cal S}_4} = 
     \mathbf{L}(\gamma) \mathbf{K}_3 \mathbf{L}(\beta)
     \mathbf{K}_2 \mathbf{L}(\alpha),
     \label{eq:transf_st1_st4}
\end{equation}
where $\mathbf{K}_i\in\mathcal{SE}(3)=\mathbf{H}^{{\cal S}_i,\widetilde{\cal S}_i} \left.\mathbf{G}^{{\cal S}_i,\widetilde{\cal S}_i}\right.^{-1}$ (similar to \eqref{eq:trans32}).

Now, if we have an additional correspondence between ${\cal S}_1$ and ${\cal S}_4$ (let's say $\mathbf{p}_{7}$), we write
\begin{equation}
\label{eq:4cam_cons1}
    \mathbf{p}_{7}^{\widetilde{\cal S}_4} \simeq
    \mathbf{L}(\gamma) \mathbf{K}_3 \mathbf{L}(\beta)
     \mathbf{K}_2 \mathbf{L}(\alpha)\mathbf{p}_{7}^{\widetilde{\cal S}_1}.
\end{equation}
Notice that we have three equation and three unknowns, meaning that in general one can get a solution for the relative poses using a single point correspondence.

To solve the problem, we take the three constraints in \eqref{eq:4cam_cons1}, together with $c\alpha^2 + s\alpha^2 = 1$, $c\beta^2 + s\beta^2 = 1$, and $c\gamma^2 + s\gamma^2 = 1$.
Since in this case we have many unknowns and high degree polynomial equations, we aim at using automatic solvers (e.g. \cite{kukelova12,larsson17}). 
In this paper we use the automatic {\it Grobner Basis} generator provided in \cite{kukelova08}.
As inputs for the automatic generator, we give the unknowns $c\alpha$, $c\beta$, $c\gamma$, $s\alpha$, $s\beta$, \& $s\gamma$ and the three constraints of \eqref{eq:4cam_cons1} plus the three trigonometric constraints.
The solver gives up to 16 solutions, as indicated in the Tab.~\ref{tab:sumup3D}.

\subsection{5-Cycle Registration}
\label{sec:567solvers}

We start by trying a general method for $n-$cycles, and show that is feasible only till $n=5$. Similar to the cases defined in the previous subsections, we consider two point correspondences between the sequences of 3D scans (without closing any cycle).
Using this data and considering the previously defined predefined transformations (Sec.~\ref{sec:get_dof}), we get matrices $\mathbf{K}_i$ as shown in \eqref{eq:trans32} and \eqref{eq:transf_st1_st4}. Using this information and applying the predefined transformations to the first and last point-clouds (similar to \eqref{eq:trans_predef_3cycle}), for an $n-$cycle loop we define the transformation from $\widetilde{\mathcal{S}}_1$ to $\widetilde{\mathcal{S}}_n$ as
\begin{equation}
\label{eq:bigfour_trans}
    \mathbf{T}^{\widetilde{\cal S}_1,\widetilde{\cal S}_n} =
    \mathbf{L}(\theta_{n-1}) \mathbf{K}_{n-1} \mathbf{L}(\theta_{n-2}) \cdots \mathbf{L}(\theta_2) \mathbf{K}_2 \mathbf{L}(\theta_1),
\end{equation}
where $\theta_i$ are the unknown degrees of freedom.

Now, for any $n = \{5,6,7\}$, we will have between four to six degrees of remaining unknowns. Since each point correspondence between the first and the last 3D scans generates three constraints, we will need two point correspondences to close the loop between $\widetilde{\cal S}_1$ and $\widetilde{\cal S}_n$:
\begin{equation}
\label{eq:other_constraints}
    \mathbf{p}_{l+1}^{\widetilde{\cal S}_n} \simeq
    \mathbf{T}^{\widetilde{\cal S}_1,\widetilde{\cal S}_n}\mathbf{p}_{l+1}^{\widetilde{\cal S}_1} \ \ \text{and}\ \ \ 
    \mathbf{p}_{l+2}^{\widetilde{\cal S}_n} \simeq
    \mathbf{T}^{\widetilde{\cal S}_1,\widetilde{\cal S}_n}\mathbf{p}_{l+2}^{\widetilde{\cal S}_1},
\end{equation}
where $l = 2(n-1)$.

Similar to what we did in the previous subsection, we use the standard {\it Grobner Basis} generator~\cite{kukelova08}. Specifically, we provide the generator $c\theta_i$ and $s\theta_i$ (a total of $2(n-1)$ variables) as the unknowns, and choose $n-1$ constraints within the set of equations in \eqref{eq:other_constraints}.
The remaining $n-1$ constraints are given by the trigonometric relations $c\theta_i^2 + s\theta_i^2 = 1$. The number of solutions for the solver with $n = 5$ is 32 (as shown in Tab.~\ref{tab:sumup3D}). As we can observe, this line of research may become computationally infeasible when $n > 5$~\cite{camposeco18,sweeney15}. For example, in the case of $n=6$, we may have up to 288 solutions and there is no easy way to build the solver.

\section{Planar Motion Case}\label{sec:planar}
We consider the problem of solving the 3D registration between scans when there is only planar motion between the point-clouds (3 degrees of freedom -- 2 translation and 1 rotation).

We note that, in Sec.~\ref{sec:get_dof}, while $\mathbf{p}_1$ is used to set the point cloud's coordinate system (see \eqref{eq:simp_pointcloud}), the $\mathbf{p}_2$ is only used to set the direction of the $z-$axis. Now, one of the features of the planar motion is that the rotation matrices between the sequences of 3D scans will have associated a single rotation angle. Without loss of generality, the respective rotation axis can be freely chosen, and in this case we choose the $z-$axis. Using this choice, one can conclude that the second point correspondence in the method presented in Sec.~\ref{sec:get_dof} is not needed. Therefore, for the computation of the predefined transformations defined in Sec.~\ref{sec:get_dof}, only one 3D point correspondence is required for each pair of 3D scans. The rest of the solvers follow the steps derived in Secs.~\ref{sec:two_camera_solver}, \ref{sec:three_camera_solver}, \ref{sec:four_camera_solver}, and \ref{sec:567solvers}.

A summary of the number of the correspondences needed for these problems, as well as the number of solutions that the solvers give is shown in Tab.~\ref{tab:sumup3DPlanar}. Notice that, in this case, the minimal solution for the two point-cloud registration is two 3D points, meaning that we are looking for cycles that consider less than two point correspondences between point-clouds. For that reason, we are only interested in mini-loop cycles up to four 3D scans.

\begin{table}
\centering
\scalebox{0.73}{\begin{tabular}{ |c|l|c|c| }
 \hline
 \bf \thead{Loop Cycle \\ \#Cameras} & \multicolumn{1}{c|}{\bf \#Correspondences}  & \bf Total & \bf \#Solutions \\
 \hline\hline
 Two   & $\#\bm{2}({\cal S}_1,{\cal S}_2)$  & 3 & 2 \\ \hline
 Three & $\#\bm{1}({\cal S}_1,{\cal S}_2); \ \#\bm{1}({\cal S}_2,{\cal S}_3) $; \ $\#\bm{1}({\cal S}_1,{\cal S}_3)$  & 3 & 4 \\ \hline
 Four  & $\begin{aligned} & \#\bm{1}({\cal S}_1,{\cal S}_2); \ \#\bm{1}({\cal S}_2,{\cal S}_3); & \\ &  \#\bm{1}({\cal S}_3,{\cal S}_4); \ \#\bm{1}({\cal S}_1,{\cal S}_4) & \end{aligned}$ & 4 & 16 \\ \hline
 \end{tabular}
 }
 \caption{\emph{This table summarizes the minimal number of correspondences required to compute the poses in $n-$cycles while considering planar motions. In the table, $\#\bm{i}({\cal S}_j,{\cal S}_k)$ means $i$ point correspondences within the sequence of point clouds ${\cal S}_j,{\cal S}_k$.}}
\label{tab:sumup3DPlanar}
\end{table}
\section{Motion Averaging}\label{sec:pose_refinement}

In this section, we show a method to use our $n-$cycle solvers to generate initial relative poses for a large collection of 3D scans. First, we construct a graph ${\cal G}=\{{\cal V},{\cal E}\}$ to denote the pose relationship between the cameras. The vertices ${\cal V}$ of this graph denote the poses of the cameras, and the edges ${\cal E}$ exist if two cameras have any scene overlap. We use SURF feature correspondences on the RGB components of the data to identify the edges for all pairs of cameras in the pose graph. We consider an edge between two cameras if we find at least $T$ feature correspondences between them.

\vspace{.25cm}
\noindent
{\bf Edge-disjoint pose graph decomposition:~}
In this method, we decompose the pose graph into edge-disjoint mini-loops. To achieve this we use a simple depth first search (DFS) traversal of the graph to identify $n-$cycles and remove the corresponding edges, so that they do not reappear in the next iteration. We first identify all the edge-disjoint 5-cycles from the graph, and then move on 4-cycles. Once we identify all the cycles with $n={3,4,5}$, the remaining edges are handled using the pairwise method. We initialize the relative poses between pairs of cameras using the associated $n-$cycle solvers, or the simple pairwise solver if an edge is not a member of an $n-$cycle.

\vspace{.25cm}
\noindent
{\bf Rotation averaging using Lie group:~}
We obtain the relative poses between different pairs of cameras using $n-$cycle minimal solvers. Due to the redundancy in the edges (i.e., we only need a set of edges in a spanning tree to uniquely compute the pose of each camera), we will have to perform some kind of averaging of the pose parameters. We use the rotation averaging framework developed by Chatterjee and Govindu~\cite{chatterjee13}. Their approach is to first consider the Lie group structure of 3D rotations and solve the rotation averaging using the $L_1$ method. Using the results from $L_1$ optimizers as initialization, they use an iteratively reweighted least squares (IRLS) approach to derive solutions that are robust to outliers. Once the rotation parameters are computed, the remaining problem is just linear in the translation and standard least squares minimization can be used.


\section{Experimental Results}\label{sec:results}
We conducted two sets of experiments: (1) 3D registration on small $n-$cycle graphs to illustrate the advantages over pairwise methods, (2) 3D registration on a large dataset by first decomposing the pose graph into smaller edge-disjoint $n-$cycles, solving the registration using minimum $n-$cycle solvers, and finally evaluating the error with respect to the ground truth. 

\begin{figure*}[t]
\centering
\begin{minipage}[t]{.25\textwidth}
\centering
\includegraphics[height=0.16\textheight]{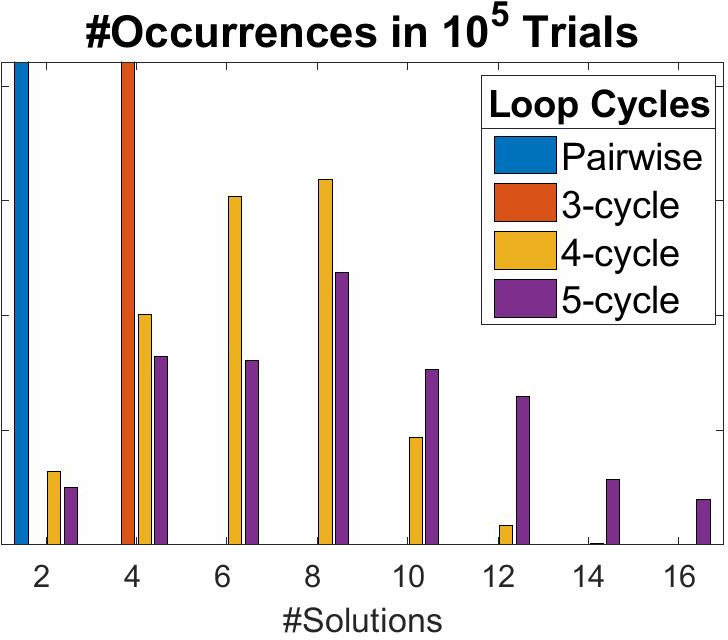}
\caption{\it Number of solutions obtained from the $n-$cycle solvers proposed in Sec.~\ref{sec:camera_solvers}. 10\textsuperscript{5} randomly generated trials were considered.}\label{fig:numOcc_compTime}
\end{minipage} \hfill 
\begin{minipage}[t]{.72\textwidth}
\centering
\includegraphics[height=0.155\textheight]{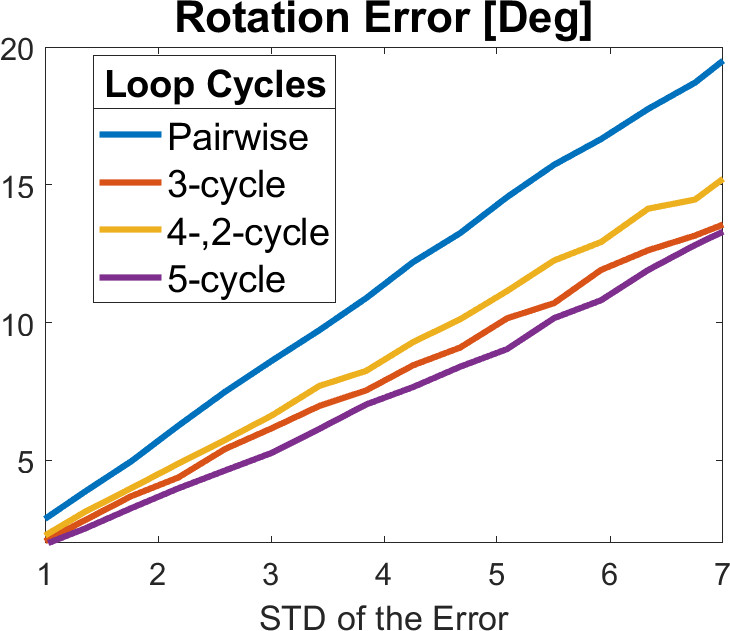}\hfill
        \includegraphics[height=0.155\textheight]{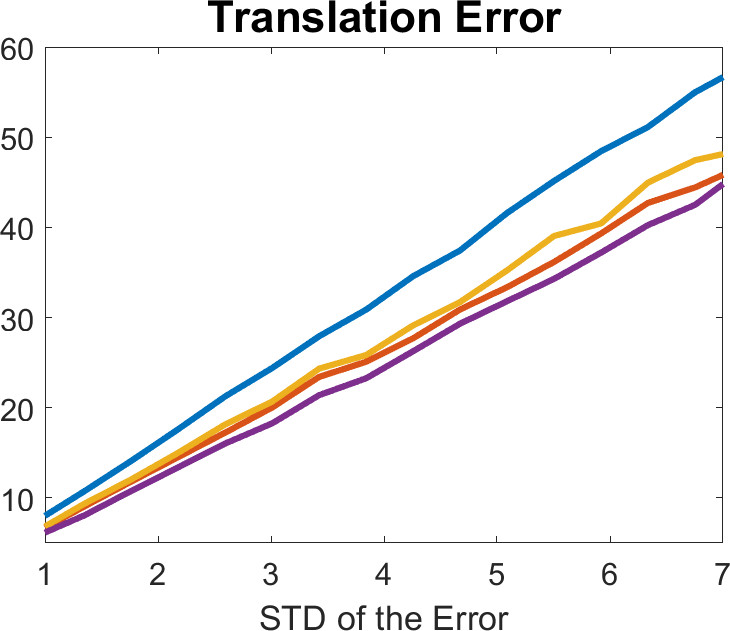}\hfill
        \includegraphics[height=0.155\textheight]{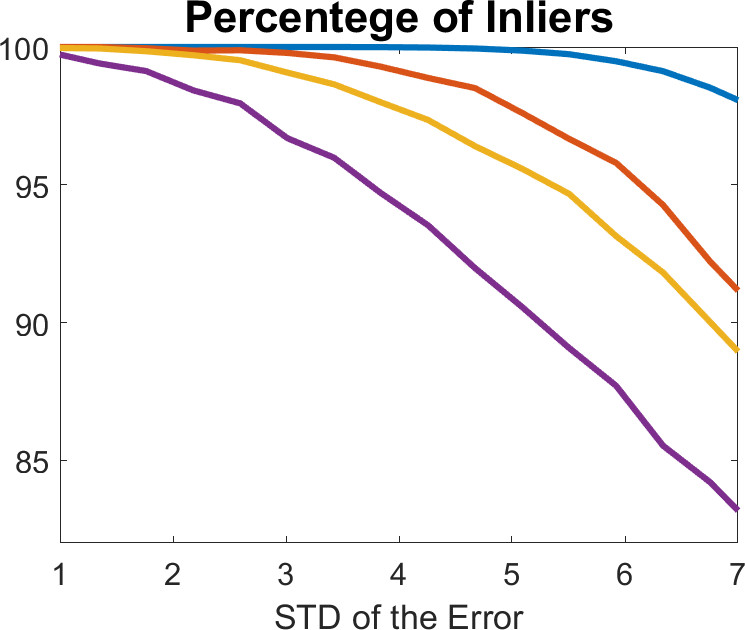}
\caption{\it
    We aim at finding the transformations between five 3D scans. We consider four different approaches: 1) {\tt Pairwise} which uses only the technique of Sec.~\ref{sec:two_camera_solver}; 2) {\tt 3-cycle} that uses only the method in Sec.~\ref{sec:three_camera_solver} (compute ${\cal S}_1$ to ${\cal S}_3$ and ${\cal S}_3$ to ${\cal S}_5$); 3) {\tt 4-,2-cycle} that uses the method in Sec.~\ref{sec:four_camera_solver} from ${\cal S}_1$ to ${\cal S}_4$ and the one in Sec.~\ref{sec:two_camera_solver} from ${\cal S}_4$ to ${\cal S}_5$; and, finally, 4) {\tt 5-cycle} that uses only the method in Sec.~\ref{sec:567solvers}.}\label{fig:ransac_eval}
\end{minipage}
\end{figure*}

\subsection{Synthetic Data}\label{sec:exp_sythetic}

We consider $400$ randomly generated 3D points and five 3D cameras in the environment, within a cube of $400$ units of side length. We consider point correspondences between different camera pairs. We select a subset of $20$ to $70\%$ random correspondences for testing our algorithms. 

\vspace{0.25cm}
\noindent{\bf Computational time and the number of solutions:~}
From the data as defined above, we select the minimal number of correspondences for each of the methods in Tab.~\ref{tab:sumup3D}, and compute the 3D registration as defined in Sec.~\ref{sec:camera_solvers}. We consider the cases: Pairwise, 3-, 4-, and 5-cycles. We repeat this procedure $10^5$ times with randomly generated data in each test. In Fig.~\ref{fig:numOcc_compTime}, we show the distribution of the number of solutions\footnote{This graphic is limited in both the number of solutions (the number of solutions for more than 16 is very small) and the number of occurrences.}. The computation time for the solvers is given in Tab.~\ref{tab:CT}. Note that the pairwise and 3-cycle cases can be computed using closed-form operations, while the 4- and 5-cycle cases require iterative techniques, this is reflected in the experimental results.

\begin{table}[t]
    \centering
    \scalebox{0.83}{\begin{tabular}{|c|c|c|c|c|}
        \hline
        {\bf Method} & 
        {\tt Pairwise} &  
        {\tt 3-cycle} & 
        {\tt 4-cycle} & 
        {\tt 5-cycle} \\ \hline \hline
        {\bf Mean}~[ms] & 
        0.0392 &  
        0.1192 & 
        3.3422 & 
        24.954 \\ \hline
    \end{tabular}}
    \caption{{\it Computation timings for $n-$cycle solvers in milliseconds (ms). Note that the implementation is in Matlab, a C++ implementation would speedup the computation time.}}
    \label{tab:CT}
\end{table}

\vspace{.25cm}
\noindent{\bf Evaluation of the proposed solvers:~}
We use Gaussian noise with a standard deviation that depends on the distance of the points from the camera center, to simulate a real 3D sensor, and the following methods:
\begin{itemize}
    \item {\tt Pairwise}: in which we use the method of Sec.~\ref{sec:two_camera_solver} to compute individual 3D registrations from ${\cal S}_1$ to ${\cal S}_2$,  ${\cal S}_2$ to ${\cal S}_3$,  ${\cal S}_3$ to ${\cal S}_4$, and ${\cal S}_4$ to ${\cal S}_5$.
    \item {\tt 3-cycle}: method in Sec.~\ref{sec:three_camera_solver} to compute transformations between ${\cal S}_1$, ${\cal S}_2$, \& ${\cal S}_3$ and ${\cal S}_3$, ${\cal S}_4$, \& ${\cal S}_5$;
    \item {\tt 4-,2-cycle}: method in Sec.~\ref{sec:four_camera_solver} to compute the 3D registrations from ${\cal S}_1$, ${\cal S}_2$, ${\cal S}_3$, \& ${\cal S}_4$, and the method in Sec.~\ref{sec:two_camera_solver} to compute the 3D registrations from ${\cal S}_4$ to ${\cal S}_5$; and
    \item {\tt 5-cycle}: method in Sec.~\ref{sec:567solvers} to compute all the transformation from ${\cal S}_1$, ${\cal S}_2$, ${\cal S}_3$, ${\cal S}_4$, and ${\cal S}_5$.
\end{itemize}
The minimal solvers were used in the RANSAC framework. A fixed number of 1000 RANSAC iterations was used, with no adaptive stopping criterion. A point distance of 50 units was used for the inlier counting. The registration from ${\cal S}_1$ to ${\cal S}_5$ is computed by multiplying each of the individual transformations from ${\cal S}_1$ to ${\cal S}_5$.

We show the angular rotation \& translation errors and the percentage inliers in Fig.~\ref{fig:ransac_eval}. 
For each level of noise, $10^3$ randomly generated trials were used.
These results show that the $n-$cycle solvers reduce the overall error in the estimation of the rotation and translation parameters.
While the $5-$cycle gives the lowest rotation and translation error, it also achieves the lowest number of inliers.

\subsection{Real Experiments}\label{sec:exp_real}

\begin{figure*}
    \centering
    \includegraphics[height=0.135\textheight]{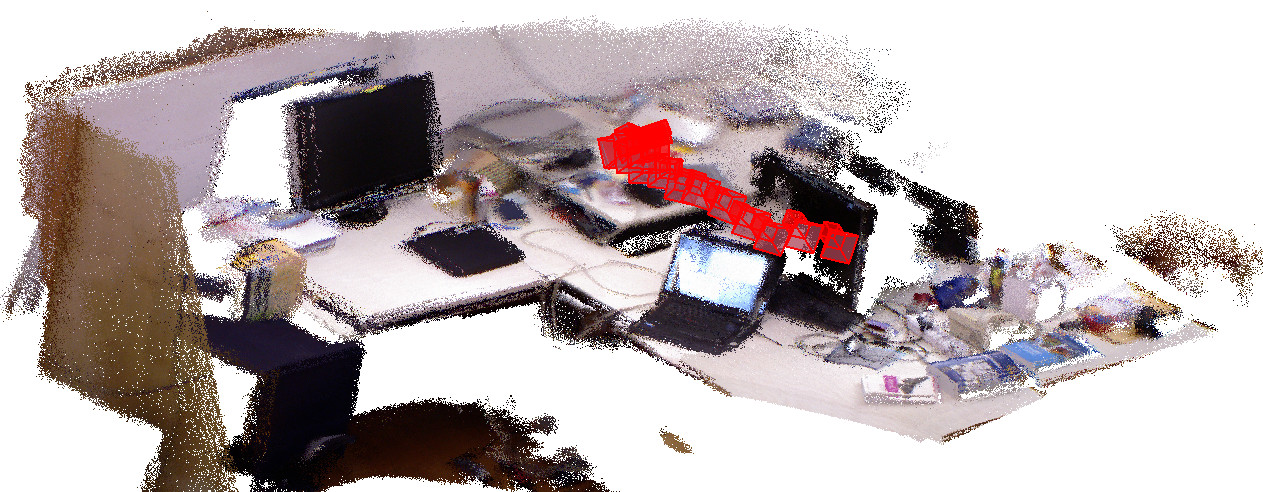}
    \includegraphics[height=0.135\textheight]{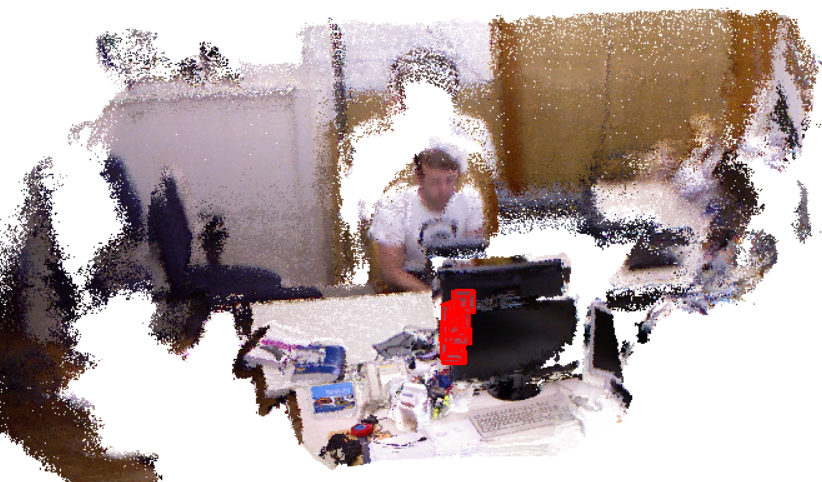}
    \includegraphics[height=0.135\textheight]{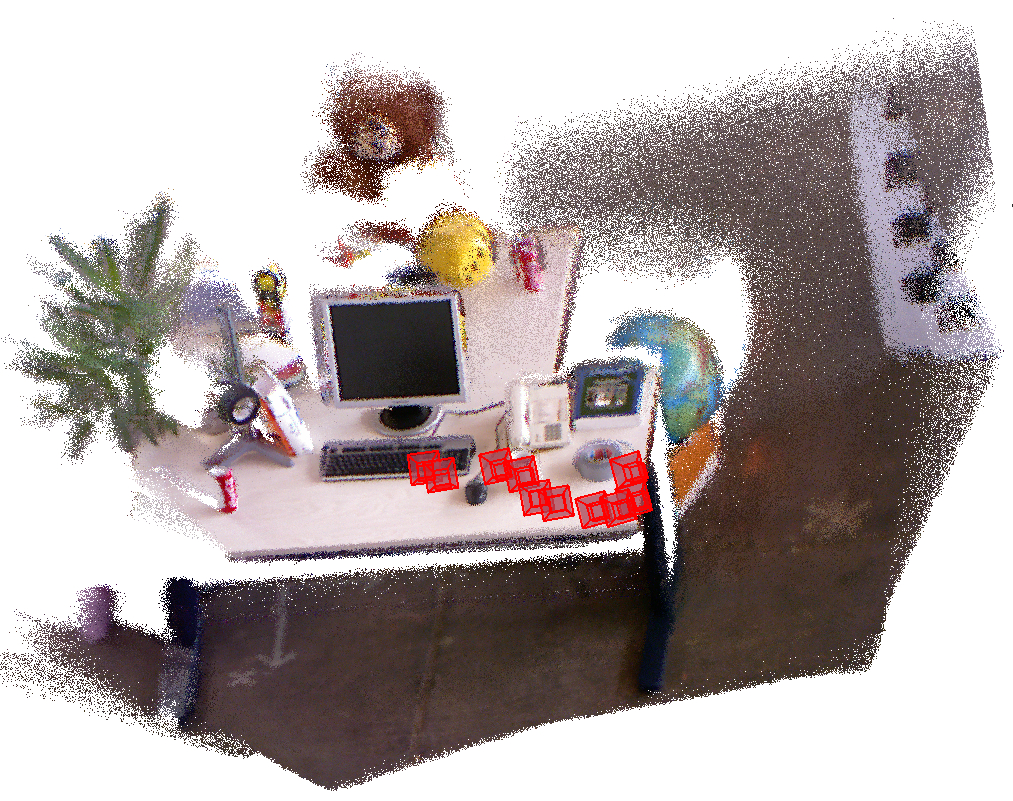}
    \caption{\emph{Results for the 3D point-cloud registration, using the TUM RGB-D data-set~\cite{tum18}. We use three different sequences of 3D scans, {\tt freiburg1\_room} (at the left),  {\tt freiburg1\_xyz} (at the center), and {\tt freiburg2\_desk} (at the right) and the method described in the paper to compute the relative transformations between the cameras. In these figures we show the registration of 100 RGB-D scans.}}
    \label{fig:final}
\end{figure*}

For real experiments, we use three sequences from the TUM dataset~\cite{tum18} that come with the ground-truth positions of the cameras ({\tt freiburg1\_room},  {\tt freiburg1\_xyz}, and {\tt freiburg2\_desk} sequences). We extract and match features using SURF~\cite{bay06} on the RGB images, and get the associated 3D points from the correspondent points in the Depth image. We start by analyzing the performance of the individual solvers separately and, then, we show their application in a large sequence using the pose graph and motion averaging discussed in Sec.~\ref{sec:pose_refinement}.

\vspace{.25cm}
\noindent{\bf Performance of the minimal solvers:~}
From the dataset, we get sequences of 5 scans with loop cycles (i.e. sets of scans with enough correspondences between $\mathcal{S}_i$ \& $\mathcal{S}_j$, to compute the poses using the respective minimal solvers). For each set of 5 scans, we compute the 3D registrations of all the 5 scans using the {\tt Pairwise}, {\tt 3-cycle}, {\tt 4-,2-cycle}, and {\tt 5-cycle} methods in a RANSAC framework, similar to what was done in the evaluation of the proposed solvers in the previous subsection. A fixed number of 2000 RANSAC iterations was used for all the methods, with no adaptive stopping criterion. A point distance of 10[cm] was used as the threshold for the inlier counting. After getting the solutions, the inliers from all the different four alternatives are injected in a non-minimal pairwise 3D registration refinement method~\cite{schonemann66}, to compute the cameras' relative position from ${\cal S}_1$ to ${\cal S}_2$, ${\cal S}_2$ to ${\cal S}_3$, ${\cal S}_3$ to ${\cal S}_4$, and ${\cal S}_4$ to ${\cal S}_5$.

The rotation and translation errors in the transformation from $\mathcal{S}_1$ to $\mathcal{S}_5$ (given by multiplying each of the pairwise transformations matrices from ${\cal S}_1$ to ${\cal S}_5$) are shown in Tab.~\ref{tab:real_errors}. The $n$-cycle methods generally outperforms the pairwise technique. The {\tt 3-cycle} performs slightly better than the {\tt 4-,2-cycle}. The {\tt 5-cycle} solver produces better results in terms of rotation errors. In addition, in Tab.~\ref{tab:real_errors} we also show the number of times that the $n-$cycles outperforms the Pairwise technique.


\begin{table}[t]
    \centering
    \scalebox{0.78}{\begin{tabular}{|c|c|c|c|c|c|c|}
        \cline{2-7}
        \multicolumn{1}{c|}{} & \multicolumn{2}{c|} {\bf Errors} & \multicolumn{2}{c|}{\makecell{\bf $n-$cycles better \\ \bf than Pairwise}} & \multicolumn{2}{c|}{\makecell{\bf $n-$cycles equal \\ \bf to the Pairwise}} \\
        \hline
        {\bf Method} & {\bf Rot.} & {\bf Tran.} & {\bf Rot.} &  {\bf Tran.} & {\bf Rot.} & {\bf Tran.} \\\hline \hline
        {\tt Pairwise}   & 0.90 & 2.53 & --- & --- & --- & --- \\ \hline
        {\tt 3-cycle}    & 0.80 & 2.44 & 53\% & 48\% & 30\% & 32\% \\ \hline
        {\tt 4-,2-cycle} & 0.80 & 2.47 & 46\% & 36\% & 36\% & 39\% \\ \hline
        {\tt 5-cycle}    & 0.77 & 2.60 & 63\% & 46\% & 8\% & 9\% \\ \hline
    \end{tabular}
    }
    \caption[\it Mean errors for the rotation (in degrees), translation (centimeters), and the number of times that the $n-$cycles outperforms the Pairwise technique, using mini sequences of 5 3D scans in the TUM dataset.]{\it Mean errors for the rotation (in degrees), translation (centimeters), and the number of times that the $n-$cycles outperforms the Pairwise technique\footnotemark, using mini sequences of 5 3D scans in the TUM dataset.}
    \label{tab:real_errors}
\end{table}

\vspace{.25cm}
\noindent{\bf TUM sequences:~}
We get 100 3D scans from the three sequences, and define a graph according to Sec.~\ref{sec:pose_refinement}. The total number of edges in the pose graph for the sequences {\tt freiburg1\_room}, {\tt freiburg1\_xyz}, and {\tt freiburg2\_desk} are 435, 1751, and 687, respectively. The number of $n$-cycle loops generated from each pose graph is shown in Tab.~\ref{tab:results}\subref{tab:results:a}.

\footnotetext{Equal in the table means that the differences in the errors computed by the $n-$cycles and pairwise are less than $10^{-4}$[deg] and $10^{-3}$[mm].}

After getting the poses on the pose graph using the proposed solvers in the RANSAC framework, we use the rotation averaging framework~\cite{chatterjee13} to compute the final rotation matrices for all the cameras. After getting the rotations from the sequences, we get the corresponding translation parameters that satisfy the 3D point correspondences, using a standard least squares minimization method. The errors in the relative poses w.r.t. the ground-truth are shown in Tab.~\ref{tab:results}\subref{tab:results:b}. The final registered scans are shown in Fig.~\ref{fig:final}.

\begin{table}[t]
\centering
\subfloat[\it \#$n$-cycle loops in the 100 3D scans.]{
\label{tab:results:a}
\scalebox{0.83}{\begin{tabular}{|r|c|c|c|c|} \hline
    {\bf Data-Set} & {\bf Pairwise} & {\bf 3-Cycle} & {\bf 4-Cycle} & {\bf 5-Cycle}  \\ \hline \hline
    {\tt freiburg1\_room} & 58 & 1 & 1 & 74 \\ \hline
    {\tt freiburg1\_xyz} & 57 & 0 & 1 & 338 \\ \hline
    {\tt freiburg2\_desk} & 53 & 2 & 2 & 124 \\ \hline
\end{tabular}}}\newline
\subfloat[\it Errors in the estimation of the transformation parameters.]{
\label{tab:results:b}
\scalebox{0.9}{\begin{tabular}{|r|c|c|} \hline
    {\bf Data-Set} & {\bf Rotation} [deg] & {\bf Translation} [cm]  \\ \hline \hline
    {\tt freiburg1\_room} & 1.96 & 4.52 \\ \hline
    {\tt freiburg1\_xyz}  & 0.740 & 2.44 \\ \hline
    {\tt freiburg2\_desk} & 1.33 & 2.26 \\ \hline
\end{tabular}
}
}
\caption{\it Results obtained for the data-sets tested in this paper, i.e. 3D registrations shown in Fig.~\ref{fig:final}. \protect\subref{tab:results:a} shows the the number of edges covered by each of the solvers, and \protect\subref{tab:results:b} presents the average of the rotation and translation errors.}
\label{tab:results}
\end{table}           
\section{Discussion}\label{sec:discussion}

The main contribution of this paper is to show that one can jointly compute the pose of the cameras in $n-$cycles using the minimal number of point correspondences. In contrast to pairwise methods, the proposed approach uses only a fewer point correspondences. For example, computing the poses of 4 cameras in 4-cycles would only require 7 point correspondences, while the pairwise methods would require a minimum of 9 correspondences (3 between every camera pair). This may come as a surprise to many of us, since we assume that we need a minimum of 3 point correspondences for registering two scans. Actually, the 3-point relative pose solver for 3D cameras is not a minimal solution. It is only a near-minimal solution. To be precise, we actually need only $2\frac{1}{3}$ point correspondences to register two scans if we count the number of pose variables and number of constraints from point correspondences. Thus we can see that for obtaining 4 camera poses (assuming one of the cameras as the reference frame), our method only requires $3\times 2\frac{1}{3} = 7$ point matches. This implies that our $n-$cycle solvers are exactly minimal, and not near minimal ones. 

The proposed solvers provide alternate ways to obtain relative poses for pairs of cameras, in addition to standard pairwise methods, and this can be very beneficial in pose graph refinement or any motion averaging framework~\cite{chatterjee13}. We observed that it is not practically feasible to solve the $n-$cycle solver when $n > 5$. 

\section*{Acknowledgements}
This work was partially supported by the Portuguese Foundation for Science and Technology (FCT), project UID/EEA/50009/2019, National Science Foundation (NSF) grant IIS 1764071, and by the Swedish Foundation for Strategic Research (SSF), project COIN. We thank the reviewers and ACs for valuable feedback.

{\small
\bibliographystyle{ieee_fullname}
\bibliography{egbib}
}

\end{document}